\documentclass[preprint,onecolumn,amsmath,nofootinbib]{revtex4-1}

\usepackage[utf8]{inputenc}
\usepackage{graphicx, floatrow}
\usepackage[caption=false]{subfig}
\usepackage{amssymb, amsfonts, amsmath, dsfont}
\usepackage{multirow}
\usepackage{makecell}
\usepackage[T1]{fontenc}
\usepackage{floatrow}

\usepackage[]{algorithm2e}

\usepackage{hyperref}

\DeclareRobustCommand{\App}[1]{Appendix~\ref{app:#1}}
\DeclareRobustCommand{\Fig}[1]{Figure~\ref{fig:#1}}

\DeclareRobustCommand{\Eq}[1]{Equation~\ref{eq:#1}}

\DeclareRobustCommand{\Tab}[1]{Table~\ref{tab:#1}}

\def \E{\textrm{E}} 
\def \Var{\textrm{Var}} 

\def \>{\rangle} 
\def \<{\langle}

\def\cL{\mathcal{L}}
\def\cA{\mathcal{A}}
\def\cC{\mathcal{C}}

\def\bv{\textbf{v}}
\def\bh{\textbf{h}}
\def\bx{\textbf{x}}

\def\static{\textrm{static}}

\def\edss{\textrm{EDSS}}

\newcommand{\CRBM}{CRBM}
\newcommand{\CRBMs}{CRBMs}
\newcommand{\CRBMlong}{Conditional Restricted Boltzmann Machine}

\def\be{\begin{equation}} 
\def\ee{\end{equation}} 
\def\longrightharpoonup{\relbar\joinrel\rightharpoonup}
\def\longleftharpoondown{\leftharpoondown\joinrel\relbar}

\def\longrightleftharpoons{
  \mathop{
    \vcenter{
      \hbox{
      \ooalign{
        \raise1pt\hbox{$\longrightharpoonup\joinrel$}\crcr
	  \lower1pt\hbox{$\longleftharpoondown\joinrel$}
	  }
      }
    }
  }
}

\newcommand \bea {\begin{eqnarray}} 
\newcommand \eea {\end{eqnarray}}

\makeatletter
\newcommand*{\defeq}{\mathrel{\vcenter{\baselineskip0.5ex \lineskiplimit0pt
                     \hbox{\scriptsize.}\hbox{\scriptsize.}}}%
                     =}
\makeatother

\begin{document}

\title{Generating Digital Twins with Multiple Sclerosis Using Probabilistic Neural Networks}

\author{Jonathan R. Walsh}
\email{jon@unlearn.ai}
\affiliation{Unlearn.AI, Inc., San Francisco, CA}

\author{Aaron M. Smith}
\affiliation{Unlearn.AI, Inc., San Francisco, CA}

\author{Yannick Pouliot}
\affiliation{Unlearn.AI, Inc., San Francisco, CA}

\author{David Li-Bland}
\affiliation{Unlearn.AI, Inc., San Francisco, CA}

\author{Anton Loukianov}
\affiliation{Unlearn.AI, Inc., San Francisco, CA}

\author{Charles K. Fisher}
\affiliation{Unlearn.AI, Inc., San Francisco, CA}

\author{for the Multiple Sclerosis Outcome Assessments Consortium}
\thanks{Data used in the preparation of this article were obtained from the Multiple Sclerosis Outcome Assessments Consortium (MSOAC). As such, the investigators within MSOAC contributed to the design and implementation of the MSOAC Placebo database and/or provided placebo data, but did not participate in the analysis of the data or the writing of this report.}


\begin{abstract}
Multiple Sclerosis (MS) is a neurodegenerative disorder characterized by a complex set of clinical assessments.  We use an unsupervised machine learning model called a \CRBMlong{} (\CRBM{}) to learn the relationships between covariates commonly used to characterize subjects and their disease progression in MS clinical trials.  A \CRBM{} is capable of generating digital twins, which are simulated subjects having the same baseline data as actual subjects.  Digital twins allow for subject-level statistical analyses of disease progression.  The \CRBM{} is trained using data from 2395 subjects enrolled in the placebo arms of clinical trials across the three primary subtypes of MS.  We discuss how \CRBM{}s are trained and show that digital twins generated by the model are statistically indistinguishable from their actual subject counterparts along a number of measures.
\end{abstract}

\flushbottom
\maketitle

\section{Introduction}
\label{sec:introduction}

Statistical models that characterize and predict disease progression have the potential to become important tools for research and management of complex, chronic indications like Multiple Sclerosis (MS).  In the context of clinical research, for example, statistical models of disease progression could be used to simulate different study designs \cite{holford2010clinical}, to identify patients who are likely to progress more rapidly than typical patients \cite{fisher_machine_2019}, or as external comparators in early stage clinical trials \cite{carrigan2020using}. Despite the availability of Disease Modifying Therapeutics (DMTs) for MS, many clinical trials still compare experimental treatments to placebo controls \cite{lublin2001placebo, polman2008ethics, uitdehaag2011changing, zhang_evolution_2019}, which can deter patients from enrolling in those trials.  At the same time, the rich history of clinical trials in MS provides a high quality source of data about disease progression under placebo or non-DMT control conditions, creating an opportunity for statistical models to further the clinical understanding of MS and better inform the design and analysis of trials~\cite{larocca_msoac_2018}.

Ideally, a statistical model of disease progression would describe all of the clinical characteristics that are relevant for a patient with MS, how these characteristics change through time, their inter-relationships, and their variation. In more technical terms, it would describe a joint probability distribution of the relevant clinical characteristics over time. Throughout, we refer to a sample drawn from this model distribution as a {\it digital subject}. By definition, therefore, a digital subject is a computationally generated clinical trajectory with the same statistical properties as clinical trajectories from actual patients. The primary advantages of digital subjects, in comparison to data from actual patients, are that they present no risk of revealing private health information and make it possible to quickly simulate patient cohorts of any size and characteristics.

Although the ability to create digital subjects enables one to simulate cohorts of patients, there are many applications for which one would like to make predictions about a particular patient. A statistical model that is able to generate digital subjects can also be used for individual patients by generating digital subjects that have the same clinical characteristics as the patient of interest at a given date (such as the start of a clinical study). We refer to a digital subject that has been generated in order to match a particular patient as a {\it digital twin} of that patient, in analogy with the usage of the term ``digital twin'' in engineering applications~\cite{grieves_digital_2016}.  For example, if the statistical model represents disease progression on a placebo, then a digital twin provides a potential outcome -- i.e., what would likely happen to this patient if she/he were given a placebo in a clinical trial? In more technical terms, a digital twin is a sample from a joint probability distribution of the relevant clinical characteristics at future times conditioned on the values of those characteristics at a previous time.  Note that it is possible to generate multiple digital twins for any patient.

\begin{figure*}[tp!]

\includegraphics[width=6.5in]{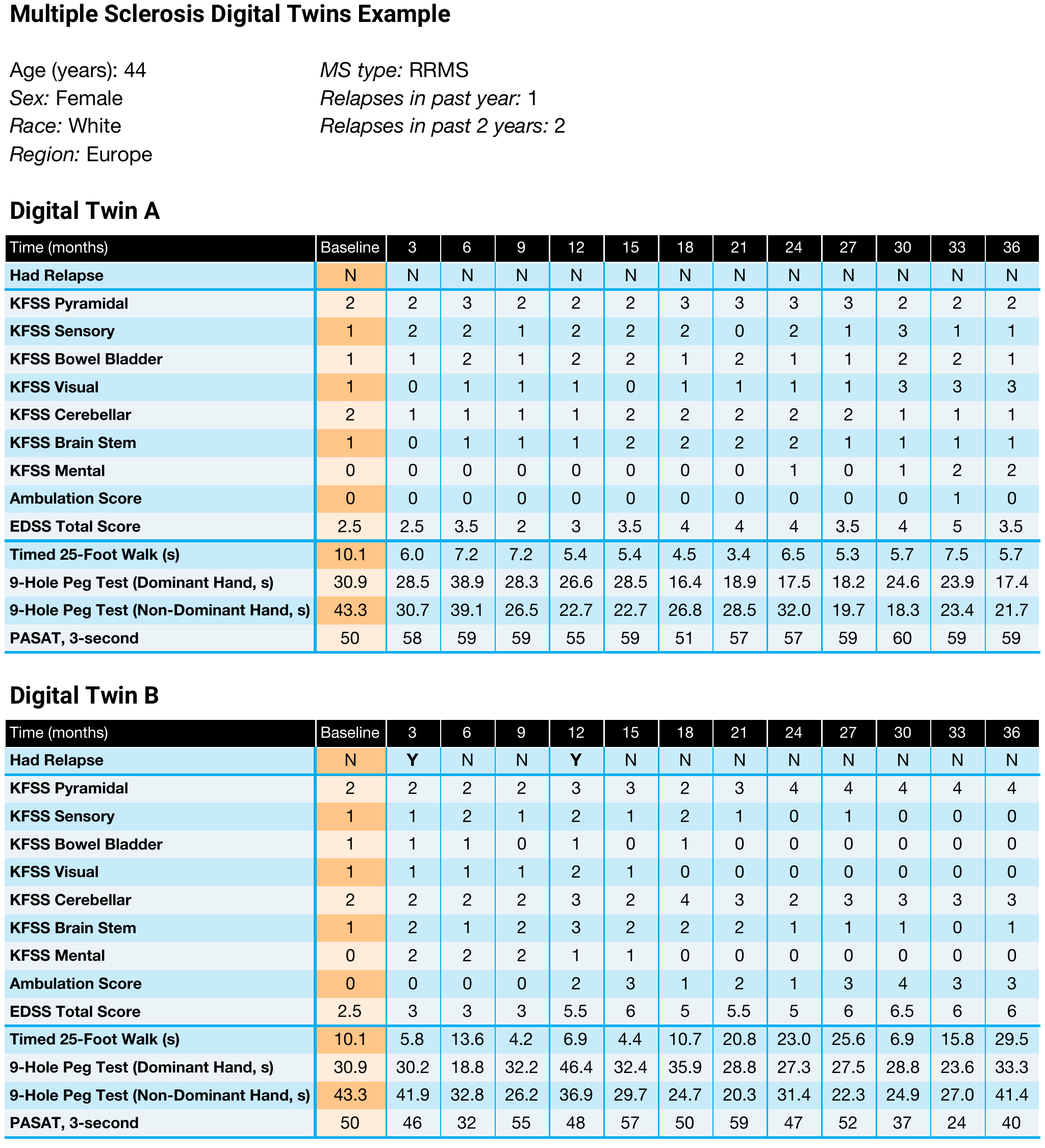}
\caption{{\bf Example digital subjects for MS.}  Two digital subjects were sampled from the joint probability distribution learned by our model so that the two subjects share the same baseline characteristics (i.e., they are digital twins). Because the model is probabilisitic, the two digital twins have different outcomes even though they share the same starting characteristics. 
\label{fig:example}}
\end{figure*}

Evaluating the quality of a model that generates digital subjects or digital twins is more complex than the evaluation of traditional machine learning models. For example, it is necessary, but not sufficient, that the means, standard deviations, correlations, and autocorrelations of all covariates computed from cohorts of digital subjects agree with those computed from cohorts of actual subjects. Here, we introduce a concept called {\it statistical indistinguishability}. Digital subjects, or digital twins, are statistically indistinguishable from actual subjects if a statistical procedure designed to classify a given clinical record as a digital subject or an actual subject performs consistently with random guessing.

Creating a statistical model that can generate digital subjects and digital twins that are statistically indistinguishable from actual patients is no small feat, but recent advances in generative neural networks make it possible to train generative models for complex probability distributions \cite{fisher_boltzmann_2018, taylor_factored_2009, taylor_modeling_2007}, including time-dependent clinical data~\cite{fisher_machine_2019}. Here, we train a probabilistic generative neural network to create digital subjects with MS using a database of placebo arms aggregated across 8 historical clinical trials~\cite{larocca_msoac_2018} covering many clinically relevant covariates including demographic information, relapses, the Expanded Disability Status Scale (EDSS)~\cite{kurtzke_rating_1983}, and functional tests. To illustrate the concept of a digital subject for MS, \Fig{example} shows two digital subjects generated by the model with the same starting characteristics (i.e., the two subjects are digital twins). 

As a complex neurodegenerative disorder with a course that is ``highly varied and unpredictable''~\cite{goldenberg_multiple_2012}, MS is both a good test-case for the application of machine learning methods as well as a disease area that would benefit from improved characterization and prediction of disease progression. Previous models for disease progression in MS include survival models~\cite{scalfari_age_2011,tremlett_new_2010,vukusic_natural_2007}, Markov models~\cite{gauthier_predicting_2007,palace_uk_2014,hou_natural_2018}, and other methods~\cite{pellegrini_predicting_2019,law_machine_2019,lawton_longitudinal_2015,achiron_longitudinal_2003,achiron_predicting_2004}.  Most of these studies aim to predict the progression of either the total EDSS score or relapses as they are common ways to characterize disease severity.  To our knowledge, our approach is the first that models comprehensive subject-level clinical trajectories that go beyond the total EDSS score.  In this case we model 20 covariates relevant to capturing MS progression (see \App{supp_data_processing}).

\section{Methods}
\label{sec:methods}

\subsection{Generating Digital Subjects with a \CRBMlong{}}
\label{sec:methods_model}

Clinical data have many properties that make statistical modeling challenging.  Data can come in varied modalities such as binary, ordinal, categorical, and continuous.  Furthermore, observations may be occasionally or frequently missing.  \CRBMlong{}s (\CRBM{}s) are well suited to address these challenges, and in past work we have found that \CRBM{}s are able to effectively model disease progression in Alzheimer's Disease~\cite{fisher_machine_2019}.

A \CRBM{}\footnote{We use a version of \CRBMs{} that have purely undirected connections, for which only a composite likelihood may be used for training, that is more appropriate for this application; see \App{crbms} for discussion.} is a class of probabilistic neural network that learns a joint probability distribution over time, and is closely related to Restricted Boltzmann Machines~\cite{ackley1985learning,hinton2006reducing,salakhutdinov2009deep, ackley1985learning, hinton2010practical, goodfellow2014generative}.  A \CRBM{} learns the relationship between covariates at $k$ visits expressed as a parametric probability distribution. For example, for $k=3$ with a 3-month spacing between visits the distribution takes the form
\begin{equation}
p( {\bf x}(t+3), {\bf x}(t), {\bf x}(t-3) ) = Z^{-1} \int d {\bf h} \, e^{-U( {\bf x}(t+3), {\bf x}(t), {\bf x}(t-3), {\bf h}) } \,,
\end{equation}
in which ${\bf x}(t)$ is the vector of covariates at time $t$ (in months), ${\bf h}$ is a vector of hidden variables, $U(\cdot)$ is called the energy function, and $Z$ is a normalization constant. More details are provided in \App{crbms}.

The \CRBM{} allows for approximate sampling from the joint probability distribution over all covariates and across $k$ visits.  Thinking of disease progression as a Markov process, the model can be used to probabilistically generate data for a subject at a visit given data from the previous $k-1$ visits.  For MS, we model data in 3-month intervals and find that modeling $k=3$ simultaneous visits in the \CRBM{} is sufficient to accurately generate long trajectories. In other words, we model MS clinical trajectories as a lag-2 Markov process.

The \CRBM{} is used in an iterative fashion to generate clinical trajectories for digital subjects or digital twins.  To generate a digital subject, the model is used to generate data at time zero (i.e., at baseline) by sampling from the marginal distribution $p( {\bf x}(t=0))$ using Markov Chain Monte Carlo (MCMC) methods. To generate a digital twin, the covariates at time zero (baseline) are set equal to the observed covariates of a particular patient. Starting from these baseline data, the \CRBM{} is used to generate data for the 3- and 6-month visits for that subject by sampling from $p( {\bf x}(t=6), {\bf x}(t=3) | {\bf x}(t=0) )$.  Then these sampled 3- and 6-month visit data may be used to generate data for the 9-month visit by sampling from $p( {\bf x}(t=9) | {\bf x}(t=6), {\bf x}(t=3) )$.  Similarly the sampled 6- and 9-month visit data may be used to generate data for the 12-month visit by sampling from $p( {\bf x}(t=12) | {\bf x}(t=9), {\bf x}(t=6) )$.  This continues as needed to create clinical trajectories of the desired length. 

Under this framework, we can now define digital subjects and digital twins more formally.  A digital subject is a clinical trajectory of length $\tau$ sampled from the joint distribution across all $\tau$ visits $\{ {\bf x}(t) \}_{t=0}^{t=\tau} \sim p( { \bf x}(t=\tau), \ldots, {\bf x}(t=0))$. A digital twin is a clinical trajectory of length $\tau$ in which the visits after $t=0$ are sampled from the conditional distribution $\{ {\bf x}(t) \}_{t=3}^{t=\tau} \sim p( {\bf x}(t=\tau), \ldots, {\bf x}(t=3) | {\bf x}(t=0))$.

One important concept is that digital subjects and digital twins are stochastic. Therefore, one can create many digital twins for a given patient. Each of these digital twins will have the same covariates at baseline, but their trajectories will differ after $t=0$. Taken together, these digital twins map out the distribution of possible clinical trajectories for that patient. Note that subjects in clinical trials do not receive different treatments at the baseline visit, so that digital twins may be created for subjects in any arm.  If the \CRBM{} models the progression of the disease under the control condition, then the model may be used to generate counterfactuals that can be used to estimate treatment effects.  In this work, our training and validation data were collected entirely from placebo control arms of previously completed clinical trials so that digital subjects generated from the model are representative of placebo controls.

\begin{figure*}[tp!]
\includegraphics[width=6.5in]{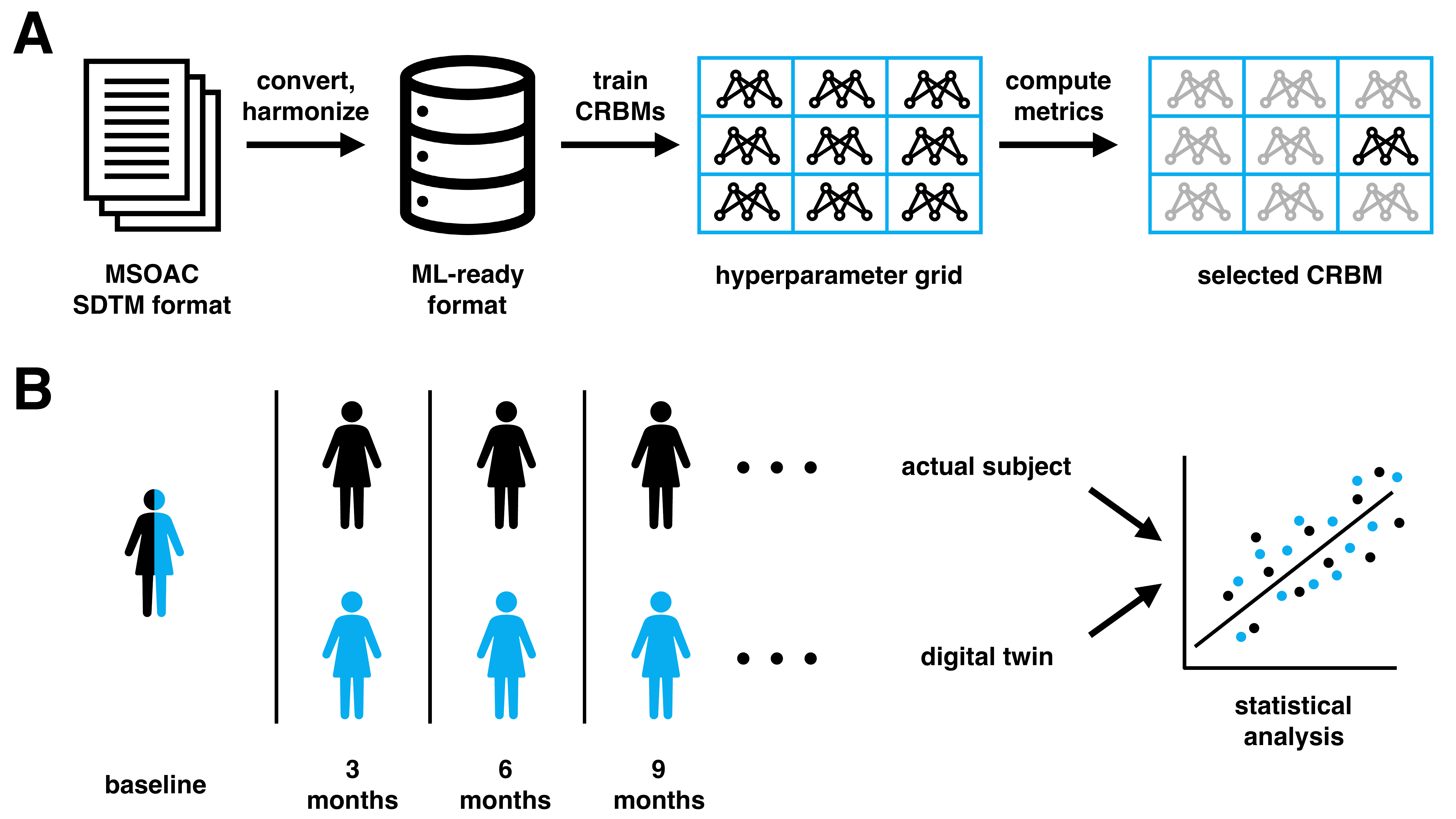}
\caption{{\bf Overview.}  A graphical summary of the process used to build the \CRBM{} studied in this paper (A), and how digital twins are analyzed (B).  To build the model, MSOAC data in SDTM format are converted to a numerical form suitable for machine learning.  A set of models are trained that sweep over a grid of hyperparameters, variations in the way the model is trained.  The final model is selected by computing metrics on validation data and choosing the best ranked model.  An important feature of the \CRBM{} is that it can create digital subjects and digital twins.  Digital subjects are synthetic clinical records generated from the model.  Digital twins are digital subjects whose baseline data is that of a given subject, allowing for subject-level predictions of outcomes.  The \CRBM{} is probabilistic, meaning many digital twins may be created for an actual subject and used to build distributions of predicted outcomes.  In this work we use digital twins to analyze the quality of the \CRBM{} as a disease progression model.
\label{fig:overview}}
\end{figure*}

A schematic overview of our process for building the \CRBM{} and generating digital twins is given in \Fig{overview}.

\subsection{Data}
\label{sec:methods_data}

One legacy of the long history of clinical development for MS is a rich set of historical clinical trials with placebo arms.  The Multiple Sclerosis Outcome Assessments Consortium (MSOAC)~\cite{larocca_msoac_2018} has compiled data from 16 clinical trials that span three subtypes of MS: relapsing remitting (RRMS), secondary progressive (SPMS), and primary progressive (PPMS). A database of placebo arms from 8 of these trials comprising 2465 subjects has been made available to qualified researchers. This database contains measurements spanning a number of domains including background information, questionnaires and functional assessments, medical history, and concomitant medications.  No imaging or biomarker data is available.

MSOAC data are encoded according to the Study Data Tabulation Model (SDTM) format, a highly structured format commonly used to submit the results of clinical trials to regulatory authorities~\cite{kubick_toward_2016,hudson_global_2018}.  Starting from the MSOAC placebo dataset, we standardized and converted SDTM-formatted measurements into a tidy format~\cite{wickham2014tidy} more suitable for statistical analysis and machine learning as described in \App{supp_data_processing}.

\begin{table*}[tp!]
\caption{Covariates used in the model are grouped into three categories: background, clinical, and functional.  Background covariates are demographics that characterize the subject population.  Functional covariates assess various aspects of MS-induced disability such as ambulation, motor skills, and cognitive function.  Clinical covariates describe a subject's clinical state such as the type of MS, degree of disability (through the Kurtzke Functional Systems Score, or KFSS, components), and relapse information.  The Statistics column gives the mean and standard deviation of each covariate (or the dominant category for binary or categorical variables) at baseline in the combined training, validation, and test sets.  The Missing column gives the percentage of missing data for each covariate at baseline.}

{\renewcommand{\arraystretch}{0.9}%
\resizebox{\textwidth}{!}{
\begin{tabular}{|c|c|c|c|c|c|}
\hline
 {\bf Name} & {\bf Category} & {\bf Type} & {\bf Longitudinal} & {\bf Statistics} & {\bf Missing [\%]} \\
\hline 
\hline
Baseline Age   &       Background   &   Continuous  &        No  & 42 (10) [years]	& 1 \\
Sex   &       Background   &   Binary    &      No  & 67\% female &	0 \\
Race    &      Background  &    Binary     &     No & 91\% white &	33 \\
Region   &       Background   &   Categorical     &     No  & 64\% Europe		& 	58 \\     
\hline
MS type     &     Clinical    &  Categorical    &      No  & 65\% RRMS 	& 0 \\   
\# of relapses, 1 year before baseline	&	Clinical	&	Ordinal	&	No	&	1.4 (0.8)	&	26	\\
\# of relapses, 2 years before baseline	&	Clinical	&	Ordinal	&	No	&	2.2 (1.3)	&	68	\\
Relapse events    &     Clinical   &   Binary     &     Yes  &  95\% no-relapse & 0 \\  
KFSS bowel and bladder system      &     Clinical   &    Ordinal     &     Yes  & 0.83 (0.98) & 13 \\
KFSS brain stem system     &      Clinical    &   Ordinal     &     Yes  &  0.64 (0.87) & 13 \\  
KFSS cerebellar system      &     Clinical    &   Ordinal    &      Yes  & 1.27 (1.17) 	& 14 \\
KFSS mental system       &    Clinical    &   Ordinal     &    Yes  & 0.58 (0.84) & 13 \\
KFSS pyramidal system    &       Clinical   &    Ordinal      &    Yes  & 1.86 (1.16)	& 13 \\ 
KFSS sensory system      &     Clinical  &     Ordinal   &       Yes  & 1.19 (1.11) & 13 \\
KFSS visual system    &       Clinical   &    Ordinal     &     Yes  & 0.76 (1.00) & 13  \\ 
Ambulation EDSS component      &     Clinical    &   Ordinal      &    Yes  & 0.72 (1.36) & 1  \\ 
\hline
Timed 25-Foot Walk       &   Functional    &  Continuous   &       Yes  & 9.5 (14.6) [s] & 1 \\
Nine-Hole Peg, dominant hand      &    Functional   &    Continuous    &      Yes  & 24.8 (15.2) [s]	& 1  \\   
Nine-hole Peg, non-dominant hand   &     Functional   &    Continuous   &       Yes  &	27.7 (22.7) [s] & 1  \\ 
Paced Auditory Serial Addition Test (3s)	&      Functional    &   Continuous   &       Yes  & 47.4 (11.8)	& 1 \\
\hline
\end{tabular}}}
\label{tab:covariates}
\end{table*}%

Twenty different covariates, described in \Tab{covariates}, were selected for inclusion in the model due to their clinical relevance and presence (non-missingness) in the dataset.  They principally span background, clinical, and functional domains.  Other clinically useful variables, such as questionnaires with data recorded for only a small fraction of subjects, were excluded.  The components of EDSS and relapse events, the primary non-imaging endpoints in clinical trials, were included in the covariates modeled.  Additionally, we added a simple longitudinal covariate that is modeled as binary, which is 1 at baseline and 0 otherwise.  This allows the model to treat baseline as a special time.

Each selected covariate was classified as either longitudinal or static and as one of binary, ordinal, categorical, or continuous; these classifications affect how the covariate is modeled by the \CRBM{}.  The resulting dataset contains clinical trajectories for all subjects in the database in 3-month intervals for up to 48 months and an average duration of approximately 24 months.  

Most subjects have a 3-month visit interval, but a 6-month interval is also present for approximately 800 subjects.  Visit days were standardized to have a baseline visit day of 0, and longitudinal measurements were grouped into windows centered on each 90-day (3-month) visit and averaged.  Any visit windows without a measurement for a particular covariate were recorded as missing.

We model each of the constituent components of the EDSS score, a combination of the 7-component Kurtzke Functional Systems Score (KFSS) assessment of function in a variety of body systems~\cite{kurtzke_evaluation_1961,kurtzke_further_1965,kurtzke_rating_1983} and the ambulatory function of the subject.  We discuss the advantages and methods around different approaches for representing these data more in \App{supp_data_processing}.

Before training, we removed 70 subjects that had essentially no data beyond baseline, resulting in a dataset with 2395 subjects.  This dataset was divided into mutually exclusive training (50\% of the data, or 1198 subjects), validation (20\% of the data, or 479 subjects), and test (30\% of the data, or 718 subjects) datasets. Due to the computational cost of associated with our training procedure, we had to rely on a single train-validation-test split, as opposed to a method like k-fold cross validation.

\subsection{Training}
\label{sec:methods_training}

A well-trained \CRBM{} is able to generate digital subjects with the same statistical properties as actual subjects in the dataset.  To accomplish this goal, we trained the \CRBM{} via stochastic gradient ascent to maximize a convex combination of a composite likelihood and an adversarial objective as previously described \cite{fisher_boltzmann_2018, fisher_machine_2019}. The relative weight given to the likelihood and adversarial components of the objective function was treated as a hyperparameter.

Hyperparameters refer to various parameters of the model or training process that must be specified, but cannot be learned by applying gradient ascent to the objective function as with the main model parameters. These hyperparameters included the weighting of the likelihood and adversarial terms of the objective function, number of epochs, batch size, initial learning rate, number of hidden units, magnitude of parameter regularization, and  sampling parameters (i.e., the number of Monte Carlo steps and the magnitude of temperature-driven sampling~\cite{fisher_boltzmann_2018}). We performed a grid search over each of these dimensions in parallel, training a total of 1296 \CRBM{}s with various hyperparameter choices to maximize an objective function on the training dataset.

After training the \CRBM{}s in the grid search, we evaluated them using a variety of metrics that assess a model's statistical performance as well as clinically important outcomes.  All metrics were computed on the validation dataset.  The statistical metrics are coefficients of determination ($R^2$ values) between the actual subjects from the test set and their digital twins the for equal-time and lagged autocorrelations. Three clinical metrics were used, one measuring the ability to predict individual relapse events, another measuring the agreement between average EDSS progression values for the data and model, and the third measuring the agreement between chronic disease worsening (CDW) fractions for the data and model.

We adopted a minimax approach to choose a best performing model across the grid search.  That is, we aimed to choose a \CRBM{} that performed well across all metrics, even if it was not the best performing model on any single metric. To accomplish this, we ranked all $N$ ($N$=1296) models from best (rank 1) to worst (rank $N$) for each metric and determined the worst rank across metrics for each model.  We used this worst rank to choose the top 25\% of models.  This selected models that performed well on all metrics. To choose a best model from this set, we repeated the process on the clinical metrics alone, ranking models on the clinical metrics and determining the worst rank of each model.  We selected the model with the lowest (best) worst rank and designated this as the ``optimal'' model.  Narrowing the focus to the clinical metrics emphasized their importance in model selection.

The hyperparameters of this optimal model were then used to train a new, ``final'' model on the combined training and validation datasets. All results that follow in the main text focus on the performance of this final model computed using the test dataset. More details on the hyperparameters, training procedures, and selection process for the optimal model are described in \App{crbms} and \App{model_selection}.

\section{Results}
\label{sec:results}

\subsection{Model Performance Across the Hyperparameter Sweep}
\label{sec:hyperparameter}

The hyperparameter sweep trained 1296 models over a grid of hyperparameters on the training dataset.  83 of these models failed to finish training due to poor performance (e.g., due to too high of a learning rate), and were excluded from further evaluation because they were ranked lowest in model selection. We computed the model selection metrics using the validation dataset on the remaining 1213 models and determined the hyperparameters of the optimal model. These hyperparameters  were used to train a new model on the combination of the training and validation datasets, which we refer to as ``the final model''. For comparison, the test metrics were computed using both the optimal model determined from the validation set and the final model trained on combining training and validation sets.

\begin{figure*}[tp!]
\includegraphics[width=6.5in]{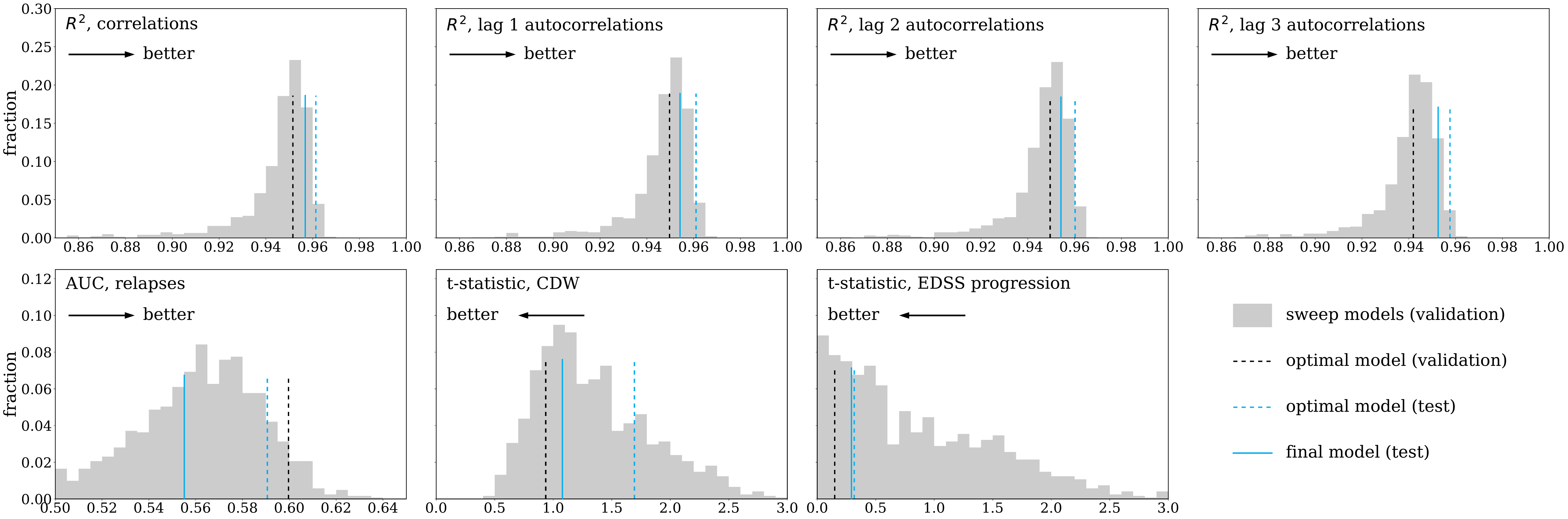}
\caption{{\bf Metrics used to select hyperparameters for the best performing model.} For each metric used, the distribution of values across the hyperparameter sweep determined using the validation dataset is shown (gray histogram).  The best performing model that was selected from this set is shown as a dashed black line at its value for each metric.  The dashed blue line shows the value of the metric for this best performing model, computed on the test dataset.  The solid blue line shows the value of the metric for the final model, computed on the test dataset.  For each metric, an arrow indicates whether smaller or larger values are better.
\label{fig:metrics}}
\end{figure*}

The distributions of the validation metrics used for the hyperparameter sweep are shown in \Fig{metrics}, along with metric values for the optimal and final models.  As expected, the optimal model performed well on all metrics without performing poorly on any individual metric.  Furthermore, the performance of the optimal model was consistent across the validation and test datasets, with the largest CDW fraction showing the largest difference. The performance of the optimal and final models on the test dataset were also quite similar, aside from the relapse prediction metric. This suggests that overall model performance is fairly stable with respect to changes to the dataset used for model training or evaluation.

\subsection{Assessing Statistical Indistinguishability of Digital Twins}
\label{sec:results_twins}

We say that digital twins and actual subjects are statistically indistinguishable if statistical analysis methods cannot differentiate the two groups better than random chance. Due to the complexity of clinical data, there are a number of ways to assess statistical indistinguishability. We focus on three different methods to quantitatively assess the degree of indistinguishability that span from subject- and covariate-level statistics to population-level statistics.  The first compares the means, standard deviations, correlations, and autocorrelations of all covariates computed from the digital twins to those computed from the actual subjects, quantifying agreement using linear regression and coefficients of determination where appropriate. The second aims to quantify the agreement between the observed clinical trajectory of a particular patient and the distribution of potential clinical trajectories defined by his/her digital twins. The third trains logistic regression models to distinguish between actual subjects and  their digital twins, quantifying performance with the area under the receiver operating characteristic curve (AUC).  

To evaluate the statistical indistinguishability of digital twins, we created 1000 digital twins for each of the 718 actual subjects in the test dataset. The statistical properties of these twins were compared to the actual subjects in a number of ways to quantify the goodness-of-fit of the \CRBM{}.

\begin{figure*}[tp!]
  \centering
  \raisebox{-0.5\height}{\includegraphics[height=3.9in]{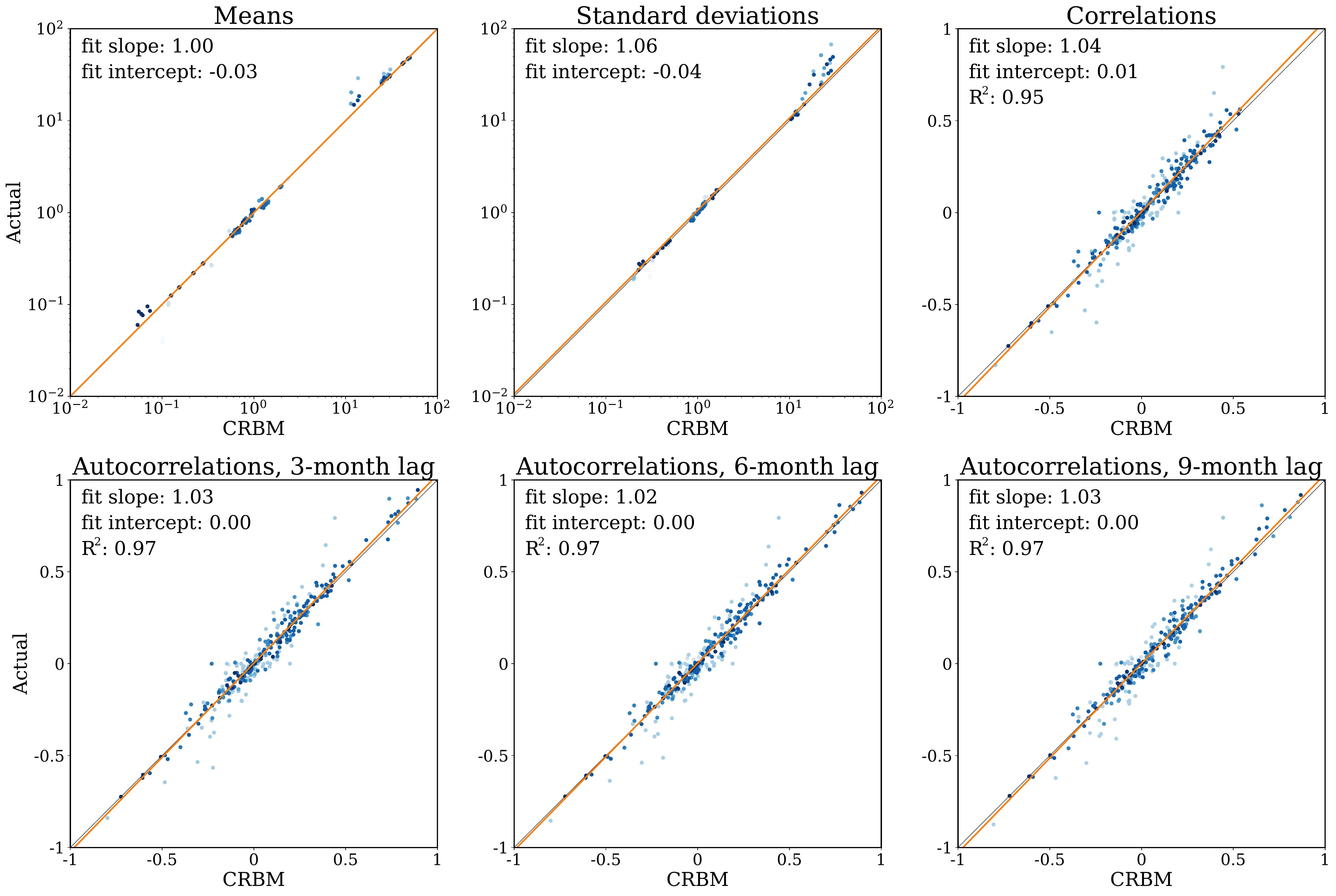}}
  \hspace*{.05in}
  \raisebox{-0.5\height}{\includegraphics[height=2in]{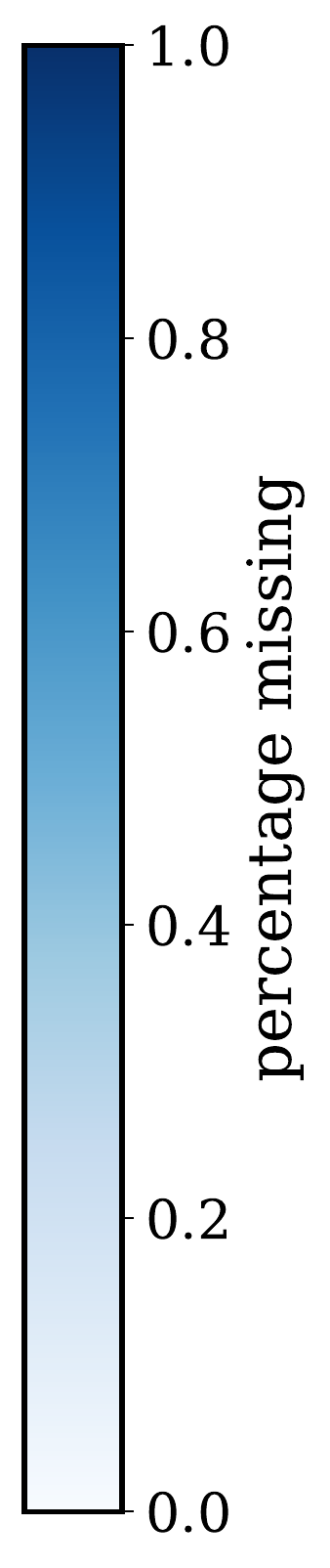}}
\caption{{\bf The model captures the leading statistical moments of the data.} Moments for the actual subject data are plotted against moments for digital twins for means, standard deviations, equal-time correlations, and lagged autocorrelations.  Each digital twin contributes data for the same study duration as its actual subject counterpart up to 36 months. The means and standard deviations are compared at each visit, with a logarithmic scale used to accommodate differing scales of the covariates.  For the equal-time and lagged autocorrelations the correlation coefficient is computed for each pair of covariates across all visits.  In all cases the slope and intercept fit coefficients shown come from regressions weighted by the fraction of data present for each covariate.  Points on the plot are darker if more data is present (see the color bar).  These regressions estimate the relationship between actual subjects and digital twins.  Theil-Sen regression is used for the means and standard deviations, which is outlier robust and appropriate for the widely varying scales present~\cite{theil_1950,sen_1968}.  For the equal-time and lagged autocorrelations, an ordinary least squares regression is used, which allows the $R^2$ values shown to be computed.  The best fit line is shown for each comparison.                                                                                                                                                                                                                                                                                                                                                                                                                                                                                                                                                                                                                                                                                                                                                                                                                                                                                                                                                                                                                                                                                                                                                                                                                                                                                                                                                                   
\label{fig:moments}}
\end{figure*}

First, in \Fig{moments} we compare the means, standard deviations, correlations, and autocorrelations between covariates computed from the actual subjects and a cohort consisting of a single digital twin for each subject. The duration of the clinical trajectory of each digital twin was the same as the matched patient. We ignored the baseline time point when computing the statistics because the actual subject and his/her digital twin have the same values of all covariates at the initial time point, by definition. All of the statistics computed from the cohort of digital twins agree with those computed from the cohort of actual subjects, with $R^2$ values for all comparisons greater than 0.95 and best fit coefficients close to their ideal values (i.e., 0 for the intercept, 1 for the slope).

\begin{figure*}
\includegraphics[width=6.5in]{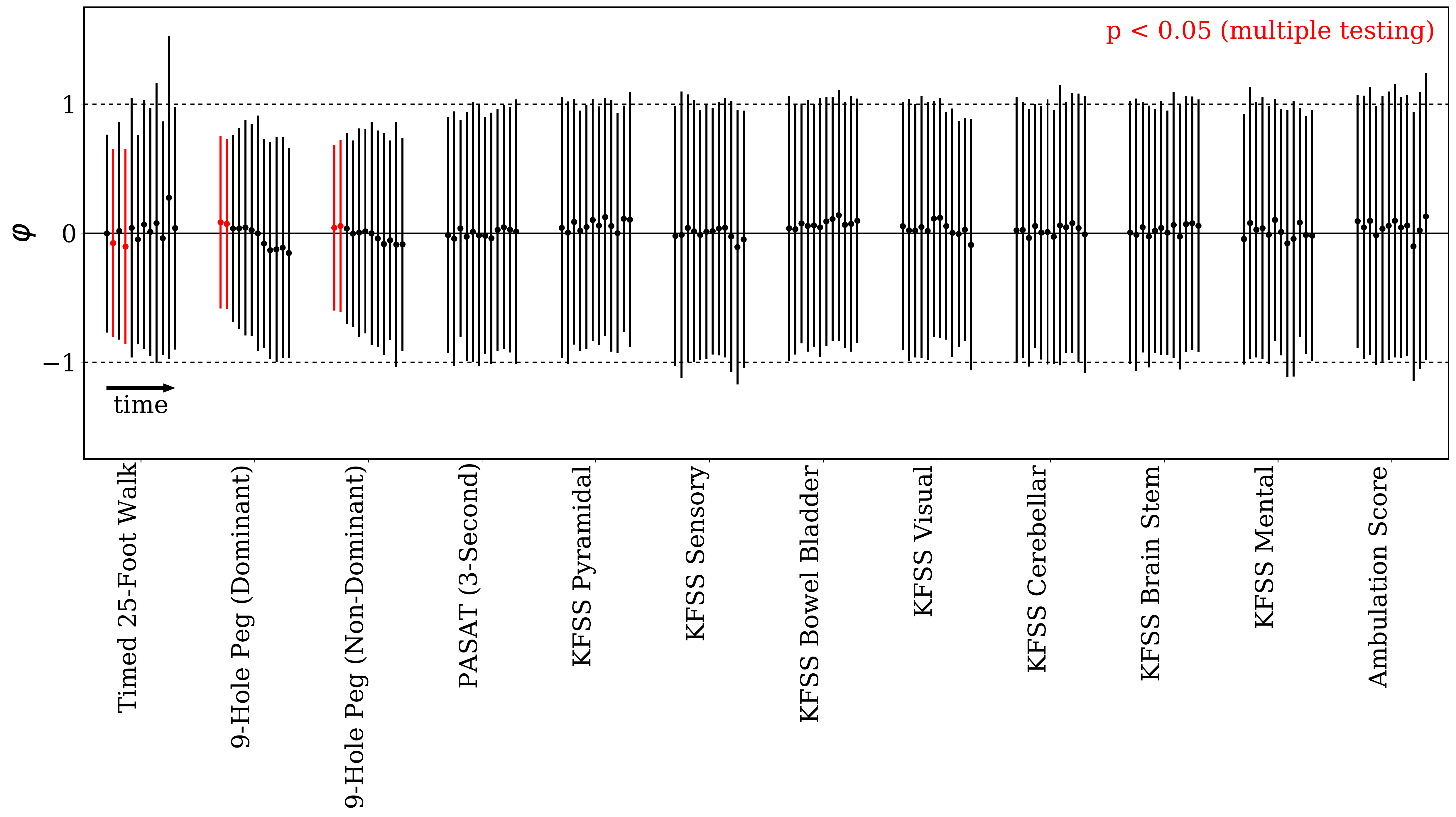}
\vspace{-0.1cm}

\floatbox[{\capbeside\thisfloatsetup{capbesideposition={left,top},capbesidewidth=4.5in}}]{figure}[\FBwidth]
{\caption{{\bf The model accurately simulates individual subject trajectories.} The mean and standard deviation of a statistic measuring subject-level agreement of the data with the \CRBM{}, for each longitudinal covariate.  Each visit beyond baseline is displayed using a point for the mean and an error bar for the standard deviation.  The visits are ordered for each covariate, with data shown up to 36 months.  For each subject, a p-value for each covariate at each visit is computed by comparing the data value to the distribution of values predicted by the \CRBM{} under repeated simulations of digital twins for that subject.  The inverse normal CDF is applied to this p-value to define a statistic, $\varphi \equiv \Phi^{-1}(p)$, and the mean and standard deviation of the distribution of $\varphi$ over subjects are computed and plotted as a point and error bar.  As the diagram on the right explains, if the \CRBM{} is in statistical agreement with the data (unbiased) then the mean should be 0 and the standard deviation should be 1.  Different types of bias are shown, and we label statistically significant cases ($p < 0.05$, with a confidence level adjusted to $0.05/144$ after applying a Bonferroni correction that factors in the 144 comparisons shown in the figure) in red.
\label{fig:phiinv_p}}}
{\includegraphics[width=1.75in]{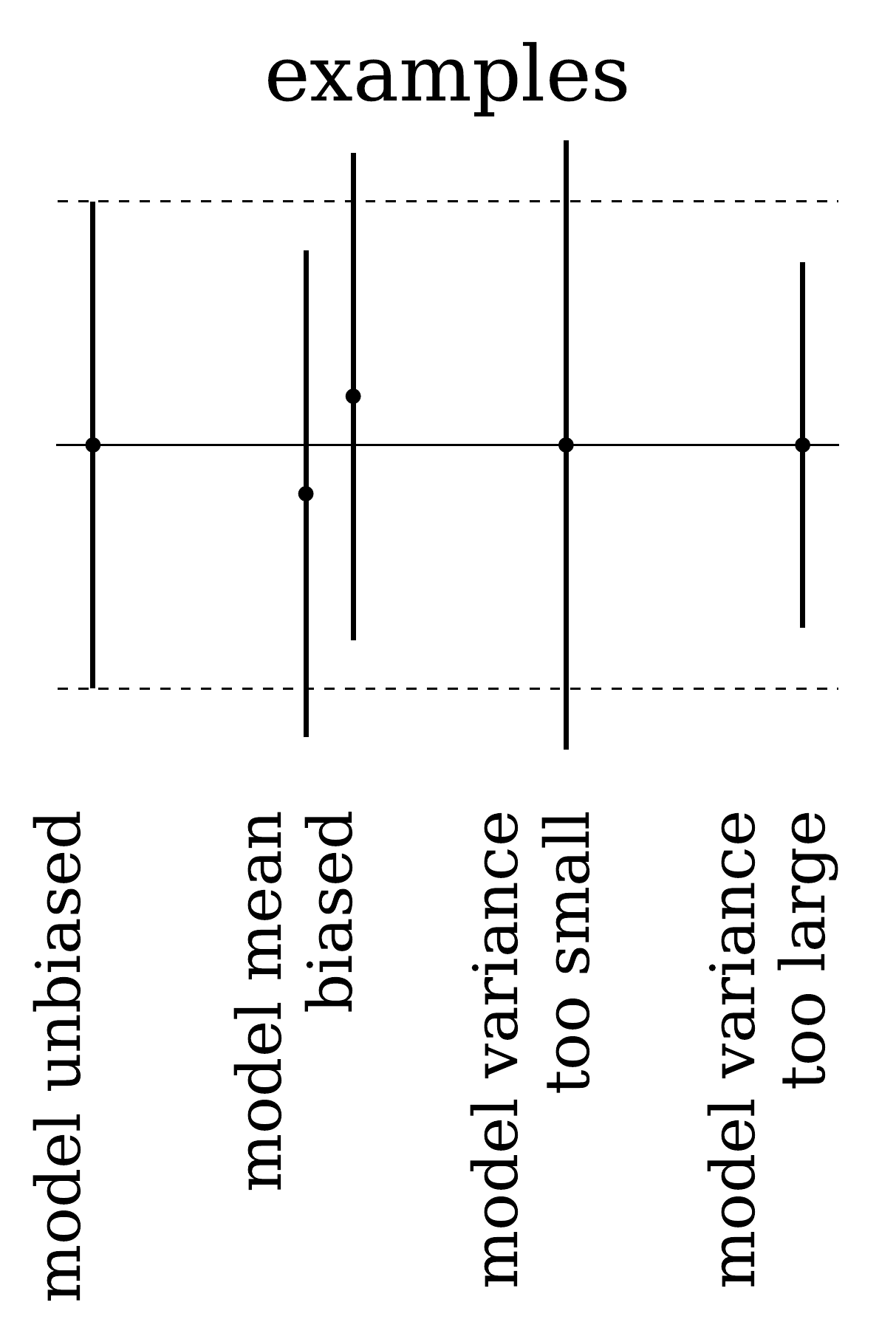}}
\end{figure*}

\Fig{moments} compares a cohort of actual subjects to a cohort of matched digital twins based on population-level statistics.  It is also important to quantify performance using statistics that are sensitive to subject-level agreement between an actual subject and his/her digital twins as in \Fig{phiinv_p}. We generated 1000 digital twins for each actual subject. For each subject, each covariate, and each visit, the samples from the digital twins form a distribution to which the observed data for the actual subject may be compared.  We used the distribution defined by the digital twins to compute a tail area probability (i.e., p-value) for each observation, then used the inverse cumulative distribution function of the standard normal distribution to convert the p-value to a statistic, $\varphi = \Phi^{-1} (p)$, that has mean 0 and standard deviation 1 if the digital twins and subject data are drawn from identical distributions.  This statistic controls for highly non-normal distributions of covariates better than a traditional z-score.  We applied the Kolmogorov-Smirnov test to assess the if the distribution of these statistics is not $\mathcal{N}(0, 1)$ and mark any cases with statistically significant disagreement at $\alpha = 0.05$ after a Bonferroni correction.

Only early visits for the timed 25-foot walk and 9-hole peg test show statistically significant differences between actual subjects their digital twins, with the difference primarily due to the standard deviation of the $\varphi$ distribution being too small.  This indicates that the standard deviations of the distributions defined for these covariates by the digital twins are too large. These two functional tests are scored as time to completion of a task. Subjects can have large single-visit outliers in this time due to short-term ambulatory or motor difficulties, which is particularly challenging to learn. Although the model generally agrees with the data well, these results suggest that the \CRBM{} generates large times for these tests too frequently.

\begin{figure*}[tp!]
\includegraphics[width=6.5in]{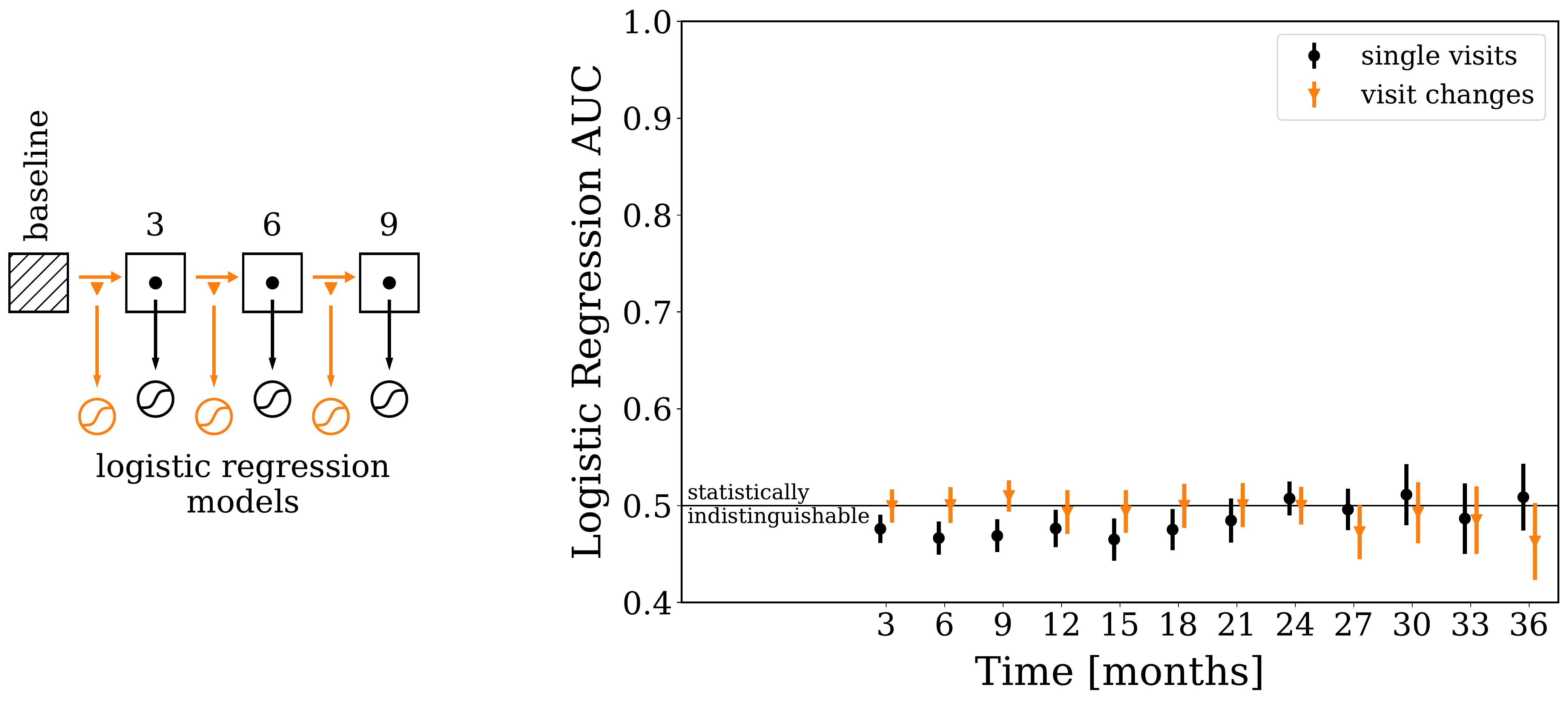}
\caption{{\bf Digital twins are challenging to distinguish from actual subjects.} The area under the receiver operating characteristic curve (AUC) of logistic regression models trained to distinguish between actual data and their digital twins are shown.  An AUC of 0.5 indicates the logistic regression model cannot differentiate between actual data and their digital twins better than random chance, while an AUC of 1 indicates the model can perfectly differentiate between the two groups.  As the diagram at left shows, at each visit beyond baseline we evaluate the ability to distinguish digital twins from data either on data from a single visit (black circles) or the change in data from one visit to the next (orange triangles).  In each case the point shown is the mean AUC is taken over 100 different simulations with different digital twins for each simulation; the error bars on the points show the standard deviation over simulations.  The AUC for each simulation is estimated using 5-fold cross validation.  Missing data from actual subjects is mean imputed; the corresponding values for the digital twins are assigned the same mean values to avoid biasing the logistic regression models.
\label{fig:adversary}}
\end{figure*}

Finally, in \Fig{adversary}, we demonstrate that a logistic regression model cannot distinguish between an actual subject and one of his/her digital twins better than random chance, which is a direct evaluation of statistical indistinguishability.  For each time point (except baseline, when the actual subjects and their twins are the same by definition), we trained a logistic regression model to distinguish between each subject and his/her digital twins. In addition, we trained logistic regression models to distinguish between each subject and his/her digital twins using differences between time points. Over 36 months, nearly every logistic regression model's performance is consistent with random guessing, meaning they are unable to distinguish between actual subjects and their digital twins better than random chance.

\section{Discussion}
\label{sec:discussion}

This work introduced three terms: {\it digital subjects}, {\it digital twins}, and the concept that digital subjects may be {\it statistically indistinguishable} from actual subjects. Using a dataset of subjects enrolled in the placebo arms of MS clinical trials, we trained a \CRBMlong{} to generate digital subjects.  This dataset included demographic information and disease history, functional assessments, and components of the EDSS score. The model learned the relationship between covariates across multiple visits, treating disease progression as a Markov process; this means that given data for a subject at some number of prior visits, the model is capable of generating clinical trajectories representing the distribution of potential outcomes at future visits for that subject. Using a variety of statistical tests, we demonstrated digital subjects generated by the \CRBM{} are statistically indistinguishable from actual subjects enrolled in the placebo arms of clinical trials for MS. 

Modeling MS disease progression is challenging due to the complexity of endpoint measures and the varied course of the disease. Relapses are difficult to predict, and the EDSS score has multiple domains and depends on many components. It is likely that significant improvements could be made to the model by incorporating additional data on imaging-related endpoints.  Incorporating these additional covariates would enable generation of digital subjects with most of the covariates that are typically measured in MS clinical trials. In addition, incorporating data from subjects treated with interferons or other commonly-used DMTs would make it possible to generate cohorts of digital subjects representing broader control populations.

The ability to generate digital twins that are statistically indistinguishable from actual subjects enrolled in control arms of clinical trials is a powerful tool to understand disease progression and model clinical trials. Each digital twin represents a potential outcome of their matched patient; that is, what would likely happen to this patient if she/he were to receive a placebo in the context of a clinical trial? Models of potential outcomes are important tools for estimating treatment effects \cite{rubin2004direct, imbens2010rubin, kunzel2019metalearners}, and ability to generate digital twins representing per-subject controls opens the door to a variety of novel clinical trial analyses aimed at assessing responses to treatments for individual patients. 

Taken in-context with previous work by the authors applying \CRBMs{} to generate digital subjects for Alzheimer's Disease \cite{fisher_machine_2019}, this work suggests that approaches based on probabilistic neural networks like \CRBMs{} will be broadly applicable to generating digital subjects across multiple diseases.

\section{Acknowledgements}
\label{sec:acknowledgements}

We would like to thank Pankaj Mehta for helpful feedback on the manuscript.

\bibliography{ms_paper_main}

\pagebreak

\appendix

\section{Data Processing and Datasets
\label{app:supp_data_processing}}

Data from the MSOAC database is stored in Study Data Tabulation Model (SDTM) format \cite{kubick_toward_2016}.  The transformation of data from SDTM to a format useful for statistical analysis and machine learning requires many steps.  The MSOAC database, which aggregates data from multiple clinical trials in SDTM format, is a clean and high-quality data source.  However, because source data in the MSOAC database are from different clinical trials, particular attention must be paid to harmonization, to ensure that data for the same concept are represented the same way.  This can easily happen, for example, if different trials use different conventions to encode measurements.

In addition to harmonization, data must be converted from SDTM format to a data format that is suitable for analysis.  SDTM represents observations in terms of variables encoding broad contextual data about the observation, while statistical analyses use tabular formats in which variables have implicit definitions and values for variables are plain data types.  This tabular format is well described by the tidy data concept~\cite{wickham2014tidy}.

To carry out data processing, we have built a software library in \texttt{python} that makes it easier to harmonize data, convert between formats, and document and test the steps taken.  This toolkit is described in~\cite{fisher_machine_2019}, and has proven effective here.

\subsection{Variables Used in Training
\label{sec:supp_covariates_details}}

We selected 20 variables from the MSOAC database for further analysis (see \Tab{supp_covariates}). These particular variables were chosen because they are frequently used in clinical trials, are relevant for predicting disease progression, and did not have a significant amount of missing data in MSOAC. The MSOAC database includes data on other questionnaires (e.g., 43 variables from SF-12, BDI-II, and RAND-36) and characterization of disability in medical history, but these data were rarely reported. In addition, MSOAC includes data on medications (16 variables) but these data lack dosing time information. Due to these limitations, we focused only on the variables described in \Tab{supp_covariates}.

Although subjects were part of placebo control arms, they may have been taking medications that are part of the standard of care for treating MS or other conditions.  Some of these medications, such as those considered to be disease modifying, are known to influence the evolution or presentation of MS.  Disease modifying therapies were rarely used (5\% of subjects received them) and were likely provided as rescue therapies from relapses or adverse events.  However, as dosing time information was not provided, these variables were not included in further analyses.

The Expanded Disability Status Scale (EDSS) is commonly used to assess disease progression in MS \cite{kurtzke_rating_1983}, but the score has a non-linear nature that makes it difficult to model. EDSS scores below 5.0 represents patients who are able to walk without aid. In this regime, the EDSS score is determined from the components of the Kurtzke Functional Systems Score (KFSS) components, a 7-component test assessing function in variety of bodily systems~\cite{kurtzke_evaluation_1961,kurtzke_further_1965,kurtzke_rating_1983}.  EDSS scores of 5.0 or greater, by contrast, are principally defined by the subject's ability to walk.  Due to the nature of scoring, some values are much more likely than others, giving a very multi-modal marginal distribution of EDSS scores.  We chose to model the individual KFSS components and ambulatory impairment of subjects separately rather than model the total score directly, allowing the model to learn the simpler constitutive covariates and the relationships between them.

Unfortunately, the MSOAC database does not record the ambulatory impairment component of EDSS directly. Instead, we infer this {\it ambulation score} from the EDSS score and encode it as an ordinal variable. The ambulation score was determined using
\begin{equation}
    \text{ambulation score} =
    \begin{cases}
      0, & \text{if           $\text{EDSS} \leq 4.5$} \nonumber \\
      2 * (\text{EDSS} - 4.5), & \text{else} \nonumber
    \end{cases}
\end{equation}
Note that we are assuming that subjects with high EDSS scores have impaired ambulatory function.  KFSS scoring rubrics do not provide consistent guidelines for assigning scores above the ambulation cutoff, indicating this assumption is valid. As a cross-check of this assumption, we took the subjects whose reported EDSS score was less than or equal to 4.5, and recomputed the score from the KFSS components. Although we found disagreements between the reported and recomputed EDSS scores in approximately 10\% of cases, in a majority of these cases we were able to clearly identify an apparent mis-scoring of EDSS based on the reported KFSS component scores. As a result, we used the EDSS scores that we recomputed directly from the KFSS components in further analyses. 
 
As with many statistical models, the performance of a \CRBM{} is often improved by scaling the variables so that their values are of order 1.  For binary and categorical covariates no normalization is applied, while for ordinal covariates we scale the covariate by its maximum so that the values range between 0 and 1.  For continuous covariates we apply different normalizing transformations depending on the features of the covariate.  Details of these transformations are provided in  \Tab{supp_covariates}.

\begin{table*}[tp!]
\caption{Variables used in the model, their domain, encoding, and the normalizing transformation applied to them.  Categorical variables (which are 1-hot encoded) and binary variables have no normalizing transformation, while ordinal variables are scaled to have a maximum value of 1.  Continuous variables are normalized in different ways, but each results in a range where values are not much larger than 1.}
{\renewcommand{\arraystretch}{0.9}%
\resizebox{\textwidth}{!}{
\begin{tabular}{|c|c|c|c|}
\hline
 {\bf Name} & {\bf Encoding} & {\bf Domain} & {\bf Normalizing Transformation} \\
\hline 
\hline
Baseline Age   &    Continuous      &   $18-72$  &     Standardization  \\
Sex   &       Binary   &   $\{0, 1\}$    &      None \\
Race    &      Binary  &    $\{0, 1\}$     &     None \\
Region   &       1-hot, 3 labels   &   $\{0, 1\}$     &     None  \\
\hline
Baseline     &     Binary    &  $\{0, 1\}$    &      None \\   
\hline
MS type     &     1-hot, 3 labels    &  $\{0, 1\}$    &      None \\   
\# of relapses, 1 year before baseline	&	Ordinal	&	$\{0, 1, \ldots, 7\}$	&	Scaled by $1/7$	\\
\# of relapses, 2 years before baseline	&	Ordinal	&	$\{0, 1, \ldots, 14\}$	&	Scaled by $1/14$	\\
Relapse events    &     Binary   &   $\{0, 1\}$     &     None \\  
KFSS bowel and bladder system      &     Ordinal   &    $\{0, 1, \ldots, 6\}$     &     Scaled by $1/6$ \\
KFSS brain stem system     &     Ordinal   &    $\{0, 1, \ldots, 5\}$     &     Scaled by $1/5$ \\
KFSS cerebellar system      &     Ordinal   &    $\{0, 1, \ldots, 5\}$     &     Scaled by $1/5$ \\
KFSS mental system       &     Ordinal   &    $\{0, 1, \ldots, 5\}$     &     Scaled by $1/5$ \\
KFSS pyramidal system    &     Ordinal   &    $\{0, 1, \ldots, 6\}$     &     Scaled by $1/6$ \\
KFSS sensory system      &     Ordinal   &    $\{0, 1, \ldots, 6\}$     &     Scaled by $1/6$ \\
KFSS visual system    &     Ordinal   &    $\{0, 1, \ldots, 6\}$     &     Scaled by $1/6$ \\
Ambulation component of EDSS score      &     Ordinal   &    $\{0, 1, \ldots, 11\}$     &     Scaled by $1/11$ \\
\hline
Timed 25-Foot Walk       &   Continuous   &     $2-300$    &  Logit scaled to range  \\
Nine-Hole Peg, dominant hand      &   Continuous   &     $10-260$    &  Logit scaled to range  \\
Nine-hole Peg, non-dominant hand   &   Continuous   &     $10-260$    &  Logit scaled to range  \\
Paced Auditory Serial Addition Test	&   Continuous   &     $0-60$    &  Logit scaled to range  \\
\hline
\end{tabular}}}
\label{tab:supp_covariates}
\end{table*}%

For age, the ``standardization'' transform means we subtract the mean and divide by the standard deviation of the training set.  The functional assessment variables (timed 25-foot walk, 9-hole peg test, and paced auditory serial addition test (PASAT)) all use a logit transformation.  The functional form of these transformations is
\be \label{eq:logistic_transform}
\tilde{x} = \log \Biggl( \frac{x - (x_- + \delta)}{(x_+ + \delta) - x} \Biggr) \,,
\ee
where $x_\pm$ are upper and lower limits of the variable (the range values given in the table) and $\delta$ is a buffer to prevent the argument from being too small or large (we use $\delta=0.5$).  These transformations, like the others used, are invertible, meaning we can transform from the representation of the data generated by the \CRBM{} back to the natural representation of the data.  For the PASAT test, we choose to model the variable as continuous (rounding the value when using the model output) rather than ordinal due to the large number of possible values.

\section{\CRBMs{}}
\label{app:crbms}

\subsection{Summary of \CRBMs{}}

Restricted Boltzmann machines (RBMs)~\cite{ackley1985learning, hinton2010practical} are a well-known class of probabilistic neural networks capable of representing the complex interrelationships between variables in large datasets.  They provide several features critical to modeling clinical data (see the supplementary material in~\cite{fisher_machine_2019}).

We use a particular type of \CRBMlong{} (\CRBM{}) that provides a way to model time series data by leveraging the natural capabilities of Boltzmann machines.  In fact, a \CRBM{} is an RBM in which the organization of the visible units is particular to the time-dependence of the data.  That is, without particular temporal labels on the visible units of the \CRBM{}, the model is simply an RBM.  These models can be trained the same way as RBMs, except that the temporal nature of the data potentially requires a reorganization of the training data.

\begin{figure*}[tp!]
\includegraphics[width=3.75in]{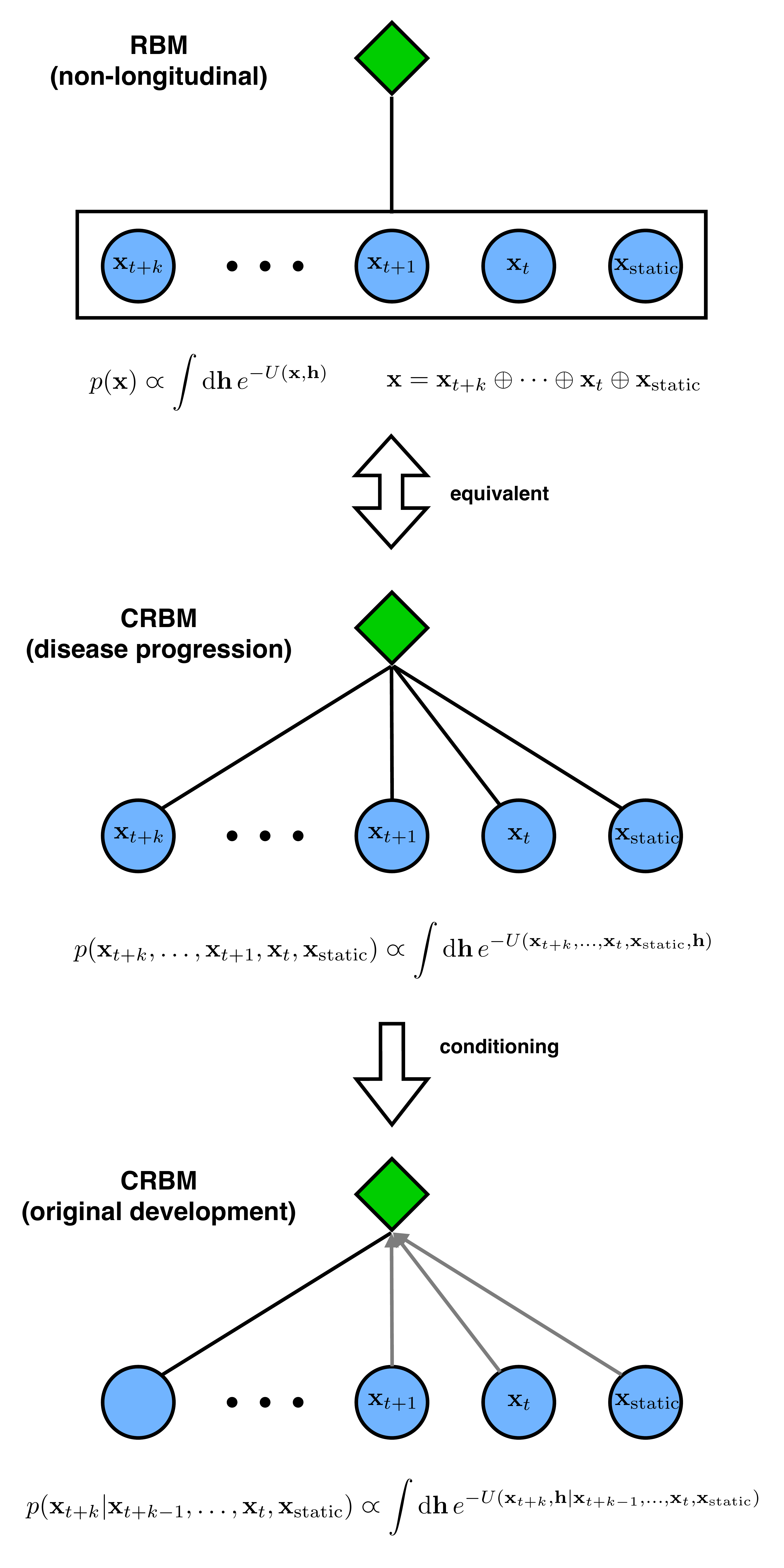}
\caption{{\bf Architectures of RBMs and \CRBMs{}.}  A schematic architecture of the \CRBM{} used in this work is shown in the middle, compared to an RBM, which is an equivalent architecture, and the original \CRBMs{}, which are described in the text.  Both the RBM and \CRBM{} used here have completely undirected connections, while the original \CRBM{} has directed connections except for the last time point.  Directed connections are shown as gray arrows.
\label{fig:crbm}}
\end{figure*}

In the original development~\cite{taylor_modeling_2007, mnih_conditional_2012}, a \CRBM{} is defined as a particular kind of graphical model encompassing an RBM whose probability distribution depends conditionally on a number ($k$) of previous time points. That is,
\begin{equation}
p( \bx_{t+k} \vert \bx_{t+k-1}, \dotsc, \bx_{t}) = Z^{-1} \int d {\bf h} \, e^{-U( \bx_{t+k}, {\bf h} \vert \bx_{t+k-1} \dotsc, \bx_{t+1}, \bx_{t}) } \,,
\end{equation}
in which $\bx_t$ is the vector of variables at time index $t$, ${\bf h}$ is a vector of hidden variables, $Z$ is a normalization constant, and $U(\cdot \vert \cdot)$ is the conditional energy function.  $K$ is the number of prior time points on which the   Here, the conditional energy function is a family of functions parametrized by linear functions of $k$ previous time points.  Graphically, this is a model in which there are directed arrows extending from the previous time points to the hidden units (and, potentially, the current visible units) of an RBM (\Fig{crbm}).

The \CRBM{} architecture described in~\cite{taylor_modeling_2007, mnih_conditional_2012} is one directional; the future is conditioned on the past. However, there are a number of reasons why we may desire a model that is bidirectional so that we could predict the future given the past or predict the past given the future. For example, the ability to predict the past given the future is especially useful for imputing missing observations -- e.g., knowledge that a subject had an EDSS score of 3.5 at their second visit provides some information on what their EDSS score likely was at the first visit, and we could use that information to impute the first visit value if it was not measured.

Our \CRBM{} architecture differs from that in~\cite{taylor_modeling_2007, mnih_conditional_2012} because it is bidirectional and it allows for some of the visible variables (such as sex or race) to be time-independent (denoted $\bx_{\mathrm{static}}$). To make the model bidirectional, we remove all connections between visible units and make all of the visible-hidden unit connections undirected (\Fig{crbm}). This leads to a joint probability distribution of $k+1$ time-adjacent vectors given by
\begin{equation}
p(\bx_{t+k}, \dotsc, \bx_{t+1}, \bx_{t}, \bx_{static}) = Z^{-1} \int d {\bf h} \, e^{-U(\bx_{t+k}, \dotsc, \bx_{t+1}, \bx_{t}, \bx_{\mathrm{static}}, {\bf h}) } \,,
\end{equation}
in which the energy function $U(\cdot)$ takes the form,
\begin{equation} \label{eq:probdist}
U(\bx_{t+k}, \dotsc, \bx_{t+1}, \bx_{t}, \bx_{\mathrm{static}}, {\bf h})  = \sum_{i={0,1,\dotsc, k,\mathrm{static}}}\biggl[\sum_{j} a_{i,j}(\bx_{t+i,j}) + \sum_{j \mu} 
W_{i,j \mu} \frac{\bx_{t+i,j}}{\sigma_{i,j}^2} \frac{h_{\mu}}{\epsilon_{\mu}^2} \biggr] + \sum_{\mu} b_{\mu} (h_{\mu}) \,.
\end{equation}
Each unit of the visible and hidden layers has bias parameters determined by the choice of functions $a_{ij}(\cdot)$ and $b_\mu(\cdot)$, as well as scale parameters $\sigma_{ij}$ and $\epsilon_\mu$.  The connection between the layers is parameterized by the weight matrices $W_{i,j\nu}$. Understood as a single RBM, our \CRBM{} contains the visible units for multiple time points coupled to a standard hidden layer.  The visible units are organized as:
\be
\bx_{\rm \CRBM{}} = \bx_{t+k} \oplus \bx_{t+k-1} \oplus \cdots \oplus \bx_{t} \oplus \bx_\static \,,
\ee
where $k$ is the time lag of the model and $\oplus$ signifies the direct sum.  The static units are only used once over all time points, as they are constant over all times.

A bidirectional \CRBM{} learns the complete joint probability distribution between all $k+1$ adjacent time points simultaneously, $p(\bx_{t+k}, \dotsc, \bx_{t}, \bx_{\mathrm{static}})$.  That means that {\it any} conditional sampling of the data may be performed, such as predicting the data for a time point given the previous $k$ time points.  In fact, the conditional distribution $p(\bx_{t+k} | \bx_{t+k-1} \dotsc, \bx_{t+1}, \bx_{t}, \bx_{static})$ has the same functional form as  that in~\cite{taylor_modeling_2007, mnih_conditional_2012}. Trajectories can be sampled by iteratively drawing from appropriate conditional distributions. For example, the trajectory $\left( \bx_{t=\tau}, \ldots, \bx_{t=1}  \right) | \left( \bx_{t=0}, \bx_{static} \right)$ is created by sampling from
\begin{align}
\bx_{t=1} &\sim p(\bx_{t=1} | \bx_{t=0}, \bx_{static}) \nonumber \\
\bx_{t=2} &\sim p(\bx_{t=2} | \bx_{t=1}, \bx_{t=0}, \bx_{static}) \nonumber \\
&\vdots \nonumber \\
\bx_{t=\tau} &\sim p(\bx_{t=\tau} | \bx_{t=\tau-1}, \ldots, \bx_{t=\tau - k}, \bx_{static}) \nonumber
\end{align}
in that sequence. Note that we can sample also from a marginal distribution like $p(\bx_{t=0}, \bx_{static})$ simply by sampling from $p(\bx_{k}, \dotsc, \bx_{0}, \bx_{\mathrm{static}})$ and discarding $(\bx_{k}, \dotsc, \bx_{1}$); therefore, it is possible to use the model both to simulate a cohort of subjects at baseline and to simulate the time-evolution of a cohort of subjects given their baseline measurements. In \Fig{crbm}, we show a schematic diagram highlighting the differences between an RBM, the  bidirectional \CRBM{} we describe here, and the directed \CRBM{} originally described in~\cite{taylor_modeling_2007,mnih_conditional_2012}.

\subsection{Training methods}

As discussed in the main text, to train the \CRBM{} we divided the 2395 subjects in the dataset up into 3 parts: training (50\% of subjects), validation (20\% of subjects), and test (30\% of subjects).  Training was done in two stages.  First we train a large number (1296) models -- each of which has a single hidden layer of ReLU units~\cite{tubiana_emergence_2017} -- over a grid of hyperparameters and measure the performance of models according to a number of metrics.  The hyperparameters of the best-performing model were used to train a model on the combination of the training and validation data.

We use a bidirectional \CRBM{} with lag $k=2$. As a result, the \CRBM{} is trained on the data from adjacent triples of time points.  If $\bx_t$ is the vector of time-dependent variables for a patient at time $t$ (in months) and $\bx_\static$ is the vector of static variables for the same patient, then the visible units used to train the \CRBM{} are $\bv = \{\bx_{t+6}, \bx_{t+3}, \bx_t, \bx_\static\}$ -- a concatenation of the data from the adjacent time points $t$, and $t+3$ months, and $t+6$ months with the static variables represented only once. Transforming the data from a subject into this triplet format involves 3 steps:
\begin{itemize}
\item[1.] For each subject we create vectors of data from all possible sets of 3 consecutive visits.  For example if a subject has data through 12 months we create vectors for 0, 3, and 6 months, 3, 6, and 9 months, and 6, 9, and 12 months.  Repeated occurrences of the static (non-longitudinal) covariates are dropped so that they only appear once in the vector.
\item[2.] Any vector that has no longitudinal data for the last visit is removed in order to remove cases in which the subject has missed visits.
\item[3.] The data are standardized according to the transformations discussed in \App{supp_data_processing}.
\end{itemize}
Each subject typically has trajectories much longer than the 6-month time window (trajectories up to 57 months are used). As a result, there are multiple triplets per subject, and the number of samples in the combined training and validation datasets (11129) is much larger than the number of subjects.  Before training, samples are randomly shuffled so that minibatches contain a mixture of subjects and times.

The \CRBM{} is trained using stochastic gradient descent to minimize an objective function that simultaneously aims to maximize the composite likelihood of the data and to minimize the ability of a classifier to discriminate between samples drawn from the model and samples taken from the training dataset \cite{fisher_boltzmann_2018}. Let $\mathcal{V}$ denote the set of all triplets of visible variables in the training set. The logarithm of the composite likelihood of the data is
\be
\mathcal{L}(\theta) \defeq \sum_{\bv \in \mathcal{V}} \log p(\bv; \theta) \, ,
\ee
in which $\theta$ denotes the parameters of the RBM. Note that two neighboring triplets from the same subject will have repeated entries and, therefore, are not independently distributed. However, we are making an approximation in which we are treating these neighboring triplets as independently distributed. Therefore, it is more appropriate to refer to $\mathcal{L}(\theta)$ as the logarithm of a composite likelihood~\cite{varin2011overview} than a true likelihood.

In \cite{fisher_boltzmann_2018}, we showed that likelihood-based approaches to training RBMs have a tendency to oversmooth the estimated distribution. To prevent this oversmoothing, we train a discriminator $q(\text{data} | \bh)$ to classify a vector of hidden unit activities as corresponding to a visible vector drawn from the data or drawn from the model. In this case, the classifier is a random forest trained on the hidden unit activities of the previous minibatch. Then, we aim to minimize an adversarial objective
\be
\mathcal{A}(\theta) \defeq - \int d \bh p(\bh; \theta) \left(2 \, q(\text{data} | \bh) - 1 \right) \, ,
\ee
which means that we aim to train the RBM so that it generates samples that the discriminator classifies as coming from the data distribution. This concept is similar to approaches used to train Generative Adversarial Networks in machine learning \cite{goodfellow2014generative}.

The final objective function $\cC(\theta)$ is a linear combination of log-composite likelihood $\cL(\theta)$ and adversarial $\cA(\theta)$ objectives,
\be \label{eq:CRBMobj}
\cC(\theta) =  -\gamma \cL(\theta) + (1-\gamma) \cA(\theta) \,,
\ee
in which $\gamma$ is a parameter -- selected as part of the hyperparameter grid search -- weighing the relative importance of the two objectives. We minimize this objective function with stochastic gradient descent using the ADAM optimizer \cite{kingma_adam_2017}.

\subsection{Temperature driven sampling}

Computing the gradients of the objective function (\Eq{CRBMobj}) requires computing various moments of the model distribution \cite{fisher_boltzmann_2018}. Unfortunately, the moments of a distribution defined by an RBM are typically intractable and can only be approximated using Monte Carlo methods. Specifically, it is possible to sample from an RBM using block Gibbs sampling by iteratively drawing $\bh \sim p(\bh | \bv)$ and $\bv \sim p(\bv | \bh)$. This creates a Markov chain that converges to the stationary distribution, but the rate of convergence may be very slow. To address this, we employ a simple approach to improve the mixing of the \CRBM{} when generating digital subjects or digital twins.  

The underlying RBM that the \CRBM{} is built upon can integrate the concept of temperature, which enters the joint probability distribution as $p_{\beta}(\bv, \bh) = Z_{\beta}^{-1} e^{-\beta E(\bv, \bh)}$, where $\beta$ is the inverse temperature and $\bv$ and $\bh$ are the visible and hidden units of the model.  By drawing the inverse temperature from an autoregressive gamma process~\cite{gourieroux2006autoregressive} with mean 1, a small standard deviation, and a non-zero autocorrelation, we can improve the rate of mixing~\cite{fisher_boltzmann_2018}.  This allows the Markov chain to mix more quickly when $\beta \ll 1$, but keeps the stationary distribution close to the case in which $\beta=1$. Unlike parallel tempering~\cite{desjardinsParallelTemperingTraining2010, desjardinsTemperedMarkovChain2010}, this algorithm for \emph{driven sampling} does not sample from the exact distribution, and instead samples from a {\it similar} distribution to the RBM that has fatter tails.  However, this approach adds only a small computation cost and appears to improve training outcomes (in this and other applications); see~\cite{fisher_boltzmann_2018} for details. 

Because varying the temperature during sampling creates some bias in the model, we use different approaches during training and testing. During training, we let the temperature vary as described above and set the autocorrelation to $0.9$ to ensure that the inverse temperature evolves slowly and the system can remain close to equilibrium. During testing, we set the autocorrelation to zero and linearly anneal the standard deviation of $\beta$ to 0 to ensure that the end of the Markov chain samples from the distribution with $\beta=1$.

\section{Hyperparameter Sweep and Model Selection}
\label{app:model_selection}

\begin{table}[hb!]
\caption{The hyperparameter grid used to train the \CRBM{}.  The grid was defined as the Cartesian product of the values listed across hyperparameters. The value of each hyperparameter for the selected model is shown in bold. The grid of batch sizes was chosen so that there were 5, 10, or 20 batches per epoch in the combined training and validation datasets. The grid of hidden units was chosen as $1/2$ or $1$ times the the number of visible units rounded up to the nearest integer. The grid of learning rates was chosen as $1/4$, $1/2$, or $1$ times the inverse of the number of visible units.}
\begin{center}
\begin{tabular}{|c|c|}
\hline
\, {\bf Hyperparameter} \, & \, {\bf Values} \, \\
\hline 
\hline
number of hidden units & {\bf 27}, 53 \\
\hline
number of epochs & 1250, {\bf 2500}, 5000, 10000 \\
\hline
batch size & {\bf 557}, 1113, 2226 \\
\hline
initial learning rate & {\bf 1/212}, 1/106, 1/53 \\
\hline
driven sampling $\sigma_\beta$ & 0, {\bf 0.15}, 0.3 \\
\hline
$\ell_2$ weight penalty & $\mathbf{10^{-4}}$, $10^{-3}$ \\
\hline
adversary weight & 0, {\bf 0.3}, 0.7 \\
\hline
Monte Carlo steps (sampling) & {\bf 100} \\
\hline
optimizer & {\bf ADAM} \\
\hline
\end{tabular}
\end{center}
\label{tab:training_params}
\end{table}%

\subsection{Hyperparameter Grid Search}

In order to train a \CRBM{}, we need to define multiple hyperparameters related to the architecture (such as the number of hidden units) and the training protocol (such as the size of the gradient steps, or learning rate). It's not clear how to choose these hyperparamaters {\it a priori}. As a result, we performed a grid search over various choices for the hyperparameters (see \Tab{training_params}), training a total of 1296 models with different hyperparameters.  Then, we selected one of these models for further analysis based on its performance on the validation dataset.

\subsection{Selecting the Best Performing Model
\label{ssec:selection_metrics}}

A \CRBM{} is a generative model that describes the probability distribution of many variables over time. Therefore, there are many potential metrics that one could use to define model performance. We addressed this by selecting a model from the hyperparameter grid search that performed well, on average, across a number of metrics.

Let $M_{ij}$ denote the rank of model $i$ on metric $j$. That is, $M_{ij} = 1$ if model $i$ is the best performing model on metric $j$, and $M_{ij} = N$ if model $i$ is the worst performing model on metric $j$. We used a two-step minimax procedure to select the best performing model on these metrics. In step one, we sorted the models based on their worst performance on any metric, $\text{MaxRank}_i = \max_j \left(M_{ij}\right)$, and discarded all models outside of the lowest quartile. In the second step, we focused on a subset $J$ of the metrics and sorted the models based on their worst performance in this subset $\text{MaxSubsetRank}_i = \max_{j \in J} \left(M_{ij}\right)$. The model with the lowest $\text{MaxSubsetRank}_i$ was selected as the best performing model overall.

This two step procedure ensured that selected model (1) did not perform poorly on any metric and (2) performed especially well on a subset of metrics. The hyperparameters of the best performing model are shown in bold in \Tab{training_params}.  These hyperparameters are used to train a model on the combination of the training and validation datasets.

Equal and time-lagged correlations between variables computed from a model that performs well should be the same as those computed from actual data. Therefore, we defined some metrics to evaluate the performance of a model based on the agreement between the lag-$l$ correlations computed from the model and those computed from the data for $l=0, 1, 2, 3$. Computing these correlations is complicated by the fact that some observations are missing. Therefore, let $x^{s}_{a,t,i}$ denote the value of variable $a$ at time $t$ in subject $i$ from source $s \in \{data, twin\}$; that is, $x^{twin}_{a,t=0,i} = x^{data}_{a, t=0, i}$. In addition, let $I_{a,t,i} = 1$ if $x^{data}_{a, t, i}$ was observed, and $I_{a,t,i} = 0$ if $x^{data}_{a, t, i}$ was missing. Then, we computed the lag-$\ell$ autocovariance between variables $a$ and $b$ from source $s$ using
\be \label{eq:autocov}
C^s_{a,b,\ell} = \frac{\sum_i \sum_t \left(x^s_{a,t,i} -  \frac{\sum_i \sum_t x^s_{a,t,i} I_{a,t,i} I_{b,t+\ell,i}}{ \sum_i \sum_t I_{a,t,i} I_{b,t+\ell,i}}\right)\left(x^s_{b,t+\ell,i}  -  \frac{\sum_i \sum_t x^s_{b,t+\ell,i} I_{a,t,i} I_{b,t+\ell,i}}{ \sum_i \sum_t I_{a,t,i} I_{b,t+\ell,i}} \right)}{ \sum_i \sum_t I_{a,t,i} I_{b,t+\ell,i} } \, .
\ee
As a metric, we defined the squared correlation coefficient between the lag-$\ell$ autocovariances computed from the data and the twins as 
\be \label{eq:Rsq_corr}
R^2_{\ell} = 1 - \frac{ \sum_{a \leq b} (C^{data}_{a,b,\ell} - C^{twin}_{a,b,\ell})^2 }{ \sum_{a \leq b} (C^{twin}_{a,b,\ell} -  \frac{1}{p (p-1)}\sum_{a \leq b} C^{twin}_{a,b,\ell})^2 } \,,
\ee
in which $p$ is the number of variables. These correlation coefficients are computed for $\ell = 0, 1, 2, 3$ ($\ell = 0$ are the equal-time and $\ell > 0$ are the lagged autocorrelations).

The metrics computed from the equal and time-lagged autocovariances capture a general notion of goodness-of-fit for the \CRBM{}, but we would also like to ensure that the model performs well on outcome measures that are frequently used as endpoints in clinical trials: the probability of a relapse at each visit, the change in the Expanded Disability Status Scale (EDSS) over 18 months, and 6-month, 1-year, and 2-year Confirmed Disability Worsening (CDW) scores. For relapses, which is one of the time-dependent variables in the model, we defined metrics in terms of the area under the receiver operating characteristic curve (AUC) that measures the model's ability to predict relapses at each visit from 3 to 18 months.

The total EDSS score is not one of the variables directly included in the model, but it can be computed because all of its components are included in the model. Let $s_i^{data}$ denote the change in the EDSS score for subject $i$ over 18 months, and let $s_{ik}^{twin}$ denote the change in the EDSS score for the $k^{th}$ digital twin of subject $i$. We defined an EDSS score metric as
\be
t_\edss = \frac{\big\lvert  \frac{1}{n} \sum_i(s_i^{data} - \frac{1}{K} \sum_k s_{ik}^{twin}) \big\rvert}{ \sqrt{ \frac{1}{n} \sum_i(s_i^{data} - \frac{1}{n} \sum_j s_{j}^{data})^2 }} \sqrt{n} \,,
\ee
in which the sums only include subjects for which the 18-month change in EDSS was observed and $n$ is the number of those subjects. We used $K=10$ digital twins for each subject. 

Similar to EDSS, CDW is not one of the variables directly included in the model but it can be computed from the model variables. A patient experiences a period of CDW if he/she has an increase of at least 1 point in EDSS (0.5 points if the initial EDSS score is greater than 6) that is sustained for (a) a continuous period of 6-months during the next year, (b) a continuous period of 3-months during the next 2 years, or (c) a continuous period of 6-months during the next 2 years. That is, there are three different definitions of CDW that we denote as CDW-a, CDW-b, and CDW-c. Let $a_i^{data}$ denote CDW-a for subject $i$, and let $a_{ik}^{twin}$ denote CDW-a for the $k^{th}$ digital twin of subject $i$. We defined a metric for CDW-a as
\be
t_{\text{CDW-a}} = \frac{\big\lvert  \frac{1}{n} \sum_i(a_i^{data} - \frac{1}{K} \sum_k a_{ik}^{twin}) \big\rvert}{ \sqrt{ \frac{1}{n} \sum_i(a_i^{data} - \frac{1}{n} \sum_j a_{j}^{data})^2 }} \sqrt{n} \,,
\ee
where the sums only include subjects with primary-progressive multiple sclerosis for whom CDW-a could be computed, and $n$ is the number of those subjects. Metrics for CDW-b and CDW-c were defined similarly. We used $K=10$ digital twins for each subject. 

All of the metrics -- those related to the autocovariances, relapses, the 18-month change in EDSS, and the CDW scores -- were used in step (1) of the model selection procedure. Only those metrics related to the relapses, the 18-month change in EDSS, and the CDW scores were used in step (2) of the model selection procedure.

\section{Model Analysis}
\label{app:model_analysis}

Figures~\ref{fig:moments},~\ref{fig:phiinv_p}, and~\ref{fig:adversary} present a number of analyses to assess the goodness-of-fit of the \CRBM{} by comparing statistics computed from the subjects in the test dataset to statistics computed from their digital twins. This section provides more details on how those analyses were performed. 

Let $x^{data}_{a, t, i}$ denote the value of variable $a$ at time $t$ that was observed for subject $i$ and let $x^{twin}_{a, t, i}$ denote the corresponding value for a digital twin sampled from the model. By definition, $x^{data}_{a, t = 0, i} = x^{twin}_{a, t = 0, i}$. In addition, let $I_{a,t,i} = 1$ if $x^{data}_{a, t, i}$ was observed, and $I_{a,t,i} = 0$ if $x^{data}_{a, t, i}$ was missing. 

Figures~\ref{fig:moments} compares the means, standard deviations, and autocovariances computed from the data to those computed from the digital twins. The means were computed as,
\be
\mu^{s}_{a,t} = \frac{\sum_i x^{s}_{a,t,i} I_{a,t,i} }{ \sum_i I_{a,t,i}} \, ,
\ee
the standard deviations were computed as,
\be
\sigma^{s}_{a,t} = \sqrt{ \frac{\sum_i (x^{s}_{a,t,i}  - \mu^{s}_{a,t,i})^2 I_{a,t,i} }{ \sum_i I_{a,t,i}} }\, ,
\ee
and the lag-$\ell$ autocovariances were computed using \Eq{autocov} for sources $s \in \{data, twin\}$. Each of the sums only includes subjects from the test data set. 

We assessed the agreement between the means and standard deviations computed from the data and those computed from their digital twins using a Theil-Sen regression. In principle, a slope of a regression line can be estimated from any pair of data points. However, this estimate for the slope is very noisy. Therefore, a Theil-Sen estimate computes many estimates for the slope from multiple pairs of data points (up to all pairs) and then uses the median slope. This procedure is robust to outliers, which is helpful here because the means and standard deviations for different variables can have different scales.

The lag-$\ell$ autocovariances can be set to the same scale by converting them to autocorrelations; i.e., by normalizing relative to the appropriate variances
\be
\rho^{s}(a,b,\ell) \defeq \frac{C^{s}(a,b,\ell)}{\sqrt{C^{s}(a,a,\ell) C^{s}(b,b,\ell)}}
\ee
for source $s \in \{ data, twin \}$. As a result, the problems that we encountered with scale differences in the means and standard deviations doesn't apply to analyzing the agreement between the autocorrelations computed from the data and those computed from their digital twins. However, each value of $\rho^{s}_{a,b,\ell}$ was only computed using pairs of data for which both variables $a$ and $b$ were observed. As a result, some autocorrelations were estimated with more precision than others. Therefore, we estimated the relationship by minimizing the weighted least-squares cost function
\be
{\text LS}_\ell \defeq \sum_{a,b} f_{a,b,\ell} (C^{data}_{a,b,\ell} - \alpha - \beta C^{twin}_{a,b,\ell})^2
\ee
in which $f_{a,b,\ell} = \sum_t \sum_i I_{a, t, i} I_{b, t+\ell. i}$. We report the estimated coefficients and the coefficient of determination in the main text. 

Although the \CRBM{} provides a model for the probability distribution of $x_{a, t}$, this distribution is generally intractable. However, we can estimate properties of this distribution using samples drawn with MCMC algorithms. Let $x^{twin}_{a, t, i, k}$ be a set of $k=1, \ldots, K$ digital twins matched to subject $i$ sampled from the \CRBM{} such that $x^{twin}_{a, t = 0, i, k} = x^{data}_{a, t = 0, i}$ for all $k$. We can use the digital twins to compute an approximate a p-value,
\be
p_{a,t,i} =  1 - \frac{1}{K} \sum_k \delta(x^{data}_{a, t, i} \leq x^{twin}_{a,t,i,k})
\ee
that describes the likelihood of observing a value more extreme than $x^{data}_{a, t, i}$ under the model distribution. Here, $\delta(\cdot)$ equals $1$ if the argument is true and $0$ otherwise. If the \CRBM{} fits the data well, then $p_{a,t,i}$ will be uniformly distributed on $(0,1)$ or, alternatively, 
\be
\varphi_{a,t, i} = \Phi^{-1} (p_{a,t, i}) \,.
\ee
will be normally distributed with mean $0$ and standard deviation $1$. Therefore, we compute the mean and variance by aggregating across subjects
\begin{align}
\E[\varphi_{a,t}] &= \frac{1}{N} \sum_i \varphi_{a,t,i} \nonumber \\ 
\Var[\varphi_{a,t}] &= \frac{1}{N} \sum_i (\varphi_{a,t,i} - \E[\varphi_{a,t}] )^2  \nonumber
\end{align}
which we plot in \Fig{phiinv_p}.

If the mean is above (below) 0, then the mean of the model is lower (higher) than the mean of the data.  If the variance is smaller (larger) than 1, then the variance of predictions of the model is larger (smaller) than the data.  Note that the relationship between the moments and the bias of the model is inverted due to the fact that the statistic is comparing the data value to the model distribution. To test significance of deviations from a standard normal distribution, we compute a Kolmogorov-Smirnov statistic between ${\cal N} (\E[\varphi_{a, t}], \Var[\varphi_{a, t}])$ and ${\cal N} (0, 1)$. We adjusted the significance level of the Kolmogorov-Smirnov test using a Bonferroni correction on the number of tests being performed, which is equal to the number of covariates evaluated (12) multiplied by the number of time points (12), for a scaling factor of 144 (meaning a comparison is significant if the p-value is less than $0.05/144$).

Finally, we introduced a concept of statistical indistinguishability in the main text by saying that digital subjects, or digital twins, are statistically indistinguishable from actual subjects if a statistical procedure designed to classify a given clinical record as a digital subject or an actual subject performs consistently with random guessing. Therefore, in order to test if the data generated by the \CRBM{} are statistically indistinguishable from the data in the test set, we should train a classifier to distinguish actual subjects from their digital twins and then assess how well it performs. 

\Fig{adversary} shows the area under the receiver operating characteristic curve (AUC) for logistic regression models trained to distinguish actual subjects from their digital twins. An AUC score of 0.5 indicates that the logistic regression cannot do better than random chance at determining whether data comes from actual subjects or their digital twins, while an AUC of 1.0 indicates that logistic regression can perfectly identify the source of a clinical record. 

Different subjects in the dataset were observed for different durations, so we trained separate logistic regression models on each visit or on the difference between consecutive visits. In each case, we computed the mean AUC taken over 100 different simulations with different digital twins for each simulation; the error bars on the points show the standard deviation over simulations.  The AUC for each simulation was estimated using 5-fold cross validation.  Missing data from actual subjects is mean imputed; the corresponding values for the digital twins are assigned the same mean values to avoid biasing the logistic regression models.


\end{document}